# Targeting Ultimate Accuracy: Face Recognition via Deep Embedding

Jingtuo Liu    Yafeng Deng    Tao Bai    Zhengping Wei    Chang Huang

Baidu Research – Institute of Deep Learning

*Abstract*—Face Recognition has been studied for many decades. As opposed to traditional hand-crafted features such as LBP and HOG, much more sophisticated features can be learned automatically by deep learning methods in a data-driven way. In this paper, we propose a two-stage approach that combines a multi-patch deep CNN and deep metric learning, which extracts low dimensional but very discriminative features for face verification and recognition. Experiments show that this method outperforms other state-of-the-art methods on LFW dataset, achieving 99.77%[1] pair-wise verification accuracy and significantly better accuracy under other two more practical protocols. This paper also discusses the importance of data size and the number of patches, showing a clear path to practical high-performance face recognition systems in real world.

1. INTRODUCTION

Recently, deep CNN based methods on face recognition problem [1, 2, 4, 6, 7, 8, 9, 12] are outperforming traditional ones with hand-crafted features and classifiers [10, 11]. The result on LFW(Labeled Faces in the Wild) [5] , a widely used dataset for evaluation of face recognition algorithms in unconstrained environment, keeps climbing as more deep CNN based methods are introduced. A common pipeline of these methods consists of two steps. Firstly, a deep CNN is used to extract a feature vector with relatively high dimension and the network can be supervised by multiclass loss and verification loss [6, 7, 8, 9]. Then, PCA [2], Joint Bayesian [6, 7, 8, 9] or metric-learning methods [12] are used to learn a more efficient low dimensional representation to distinguish faces of different identities. Some put the two stages into an end-to-end learning process [12]. Many smart methods have been used in the first step, such as joint learning [6, 8, 9], multistage feature and supervision [6, 7, 9], multi-patch features [2, 6, 7, 8, 9] and sophisticated network structure [12]. Meanwhile, huge amount of labeled face data is usually important to the performance. The amount of training data can range from 100K up to 260M.

There are discussions on how data size impacts the result of deep CNN based methods and whether the tricks are essential with different data size [2, 12]. We have investigated these issues in our experiments. According to our experiments, the quantity of faces and identities in training data is crucial to the final performance. Besides, multi-patch based feature and metric learning with triplet loss can still bring significant improvement to deep CNN result even the data size increases.

In this paper, we will introduce our two-stage method based on simple deep CNNs for multi-patch feature extraction and metric learning for reducing dimensionality. We achieve the best accuracy (99.77%) of LFW under 6000 pair evaluation protocol as well as other two protocols. Experiments will show how data size and multi-patch influence the performance. Moreover, we will demonstrate the possibility of the utilization of face recognition technique in real world as the results under other two more practical protocols are also quite promising.

2. METHOD

Our method takes two steps in training. They will be illustrated in separate sections as followed.

*2.1  Deep CNNs on Multi-patch*

We simply use a network structure with 9 convolution layers and a softmax layer at the end for supervised multi-class learning. The input of the network is a 2D aligned RGB face image. Pooling and Normalization layers are between some convolution layers. The same structure is used on overlapped image patches centered at different landmarks on face region. Each network is trained separately on GPUs. Outputs of the last convolution layer of each network are selected as the face representation and we simply concatenate them together to form a high dimensional feature.

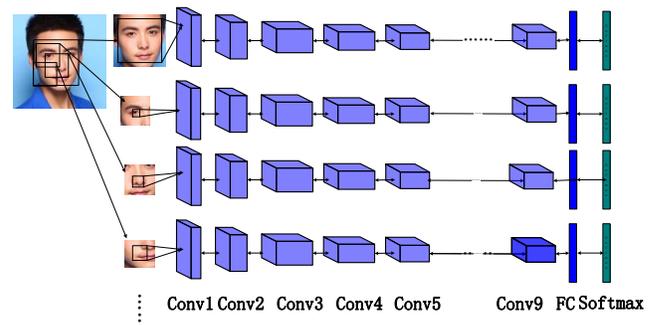

**Figure 1.**  Overview of deep CNN structure on multi-patch.

---

[1] This result has been updated from the original paper to reflect a bug fix in our ensemble algorithm for ten-fold cross validation.

## 2.2 Metric Learning

The high dimensional feature itself is representative but it's not efficient enough for face recognition and quite redundant. A metric learning method supervised by a triplet loss is used to reduce the feature to low dimension such as 128/256 float and meanwhile make it more discriminative in verification and retrieval problems. Metric learning with a triplet loss aims at shortening the L2 distance of the samples belonging to the same identity and enlarging it between samples from different ones. Hence, compared to multi-class loss function, triplet loss is more suitable for verification and retrieval problems.

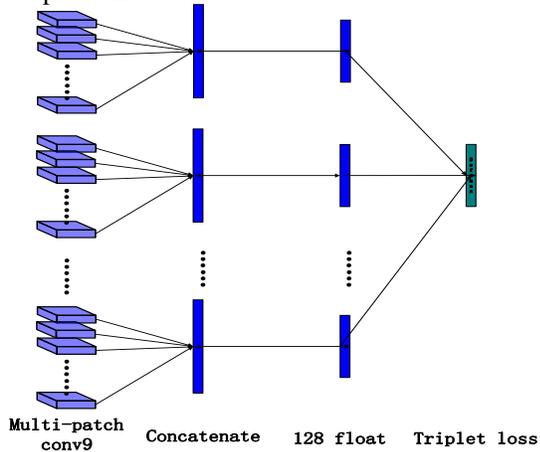

**Figure 2.** Metric learning with triplet loss

## 3. EXPERIMENTS

### 3.1 Training Datasets

We collected images of celebrities on websites, detected the faces in the image, and labeled the faces of each person by hand to remove the noises. After removing the people in LFW by name, we got about 18000 people with about 1.2 million face images. Each face is positioned and aligned by landmarks. We use the datasets to train our models.

### 3.2 Evaluation Protocols

LFW is the most popular evaluation benchmark for face recognition in real situation. There are three evaluation protocols to evaluate performances on LFW. The first protocol is to test the accuracy of 6000 face pairs, which is proposed by Huang et. al. in [5] and updated in [13], and we follow the "unrestricted, labeled outside data" task to evaluate our method. The second protocol is proposed in [3] and the protocol includes a closed-set identification task and an open-set identification task. The third protocol is proposed in [14] which include a verification task and an open-set identification task. So there are 5 tasks used to evaluate and compare our models with other methods. Please refer to [5, 3, 13, 14 ] for the details of all the protocols.

### 3.3 Data Driven

We trained three embedding models with 150K, 450K and 1.2M face images. Each of them outputs a 128-d vector as the representation of a face, and Euclidean distance in such a 128-d space properly measures the similarity of any two faces. Based on the distribution of training data, a certain threshold is estimated to tell whether two faces are of the same person or not. We tested the three models on LFW with the first protocol, and observed significant improvements with more training data as shown in Table 1: with 1.2M face images from 18K people, the 6000-pair verification error rate is reduced by more than two thirds compared to the one with 150K face image from 1.5K people.

**TABLE 1.** PAIR-WISE ERROR RATE WITH DIFFERENT AMOUNT OF TRAINING DATA

| Identities | Faces | Error rate |
|---|---|---|
| 1.5K | 150K | 3.1% |
| 9K | 450K | 1.35% |
| 18K | 1.2M | 0.87% |

### 3.4 Effect of Multi-patch

To study the effective of using multiple patches, nine CNN models were trained separately with 1.2M face images, each taking a cropped patch centered at a different landmark of certain scale. Four different embedding models are learned based on the concatenation of the outputs of one, four, seven or nine CNN models. We found this approach very effective since local patches are generally more robust to variations such as poses or expressions. As shown in Table 2, the 6000-pair error rate decreases as the number of patches increases, but somehow saturates after seven patches.

**TABLE 2.** PAIR-WISE ERROR RATE WITH DIFFERENT NUMBER OF PATCHES

| Number of patch | Error rate |
|---|---|
| 1 | 0.87% |
| 4 | 0.55% |
| 7 | 0.32% |
| 9 | 0.35% |

### 3.5 Final Performance on LFW

As described in section 3.4, the seven-patch embedding model achieves 99.68% pair-wise classification accuracy, which is already among the best published results under this protocol. Moreover, we trained another nine embedding models with different parameters, and the Euclidean distance measurements of these ten models are combined for further improvement. We simply follow the ten-fold cross-validation rule: a linear ensemble model is trained in each iteration with nine folds and tested with the leftover one. The weights of models and the classification threshold are obtained by directly minimizing the classification error of training data with a heuristic grid search algorithm. The final average testing accuracy of ten iterations is 99.77%. As for the other four tasks in second and third protocol, we simply average the distance measurements of all ten embedding models. As shown in Table 3, our single model and ensemble model both outperform all previously published results on the five tasks.

The pair-wise accuracy is the most popular protocol on LFW. The proposed approach achieves 99.77%[1] by combining ten models, which reduces the error of the previous state-of-the-art reported in [12] by about 38%. Out of all 6000 pairs, only 14 are misclassified by this ensemble model, and five of them are actually mislabeled according to latest LFW errata. The 14 misclassified pairs are listed in figure 3, as well as their scores given by the ensemble model.

Although several algorithms have achieved nearly perfect accuracy in the 6000-pair verification task, a more practical criterion for face verification applications is the false reject rate at extremely low false acceptance rate (e.g., @ 0.1%), which lies at the far end of a ROC curve rather than its central part. Moreover, open-set identification at low false acceptance rate is even more challenging but applicable to many scenarios. We also compared our single model and the ensemble model with previous published methods on these extensive protocols in Table 3. Particularly in the open-set identification task of second protocol [3], the best published identification rate is 81.4% [9], while our ensemble model can achieve 95.8% identification rate, relatively reducing the error rate by about 77%.

**TABLE 3.** COMPARISONS WITH OTHER METHODS ON SEVERAL EVALUATION TASKS

| Method | Performance on tasks | | | | |
|---|---|---|---|---|---|
| | Pair-wise Accuracy(%) | Rank-1(%) | DIR(%) @ FAR =1% | Verification(%)@ FAR=0.1% | Open-set Identification(%)@ Rank = 1,FAR = 0.1% |
| *IDL Ensemble Model* | 99.77 | 98.03 | 95.8 | 99.41 | 92.09 |
| *IDL Single Model* | 99.68 | 97.60 | 94.12 | 99.11 | 89.08 |
| *FaceNet[12]* | 99.63 | NA | NA | NA | NA |
| *DeepID3[9]* | 99.53 | 96.00 | 81.40 | NA | NA |
| *Face++[2]* | 99.50 | NA | NA | NA | NA |
| *Facebook[15]* | 98.37 | 82.5 | 61.9 | NA | NA |
| *Learning from Scratch[4]* | 97.73 | NA | NA | 80.26 | 28.90 |
| *HighDimLBP[10]* | 95.17 | NA | NA | 41.66(reported in [4]) | 18.07(reported in [4]) |

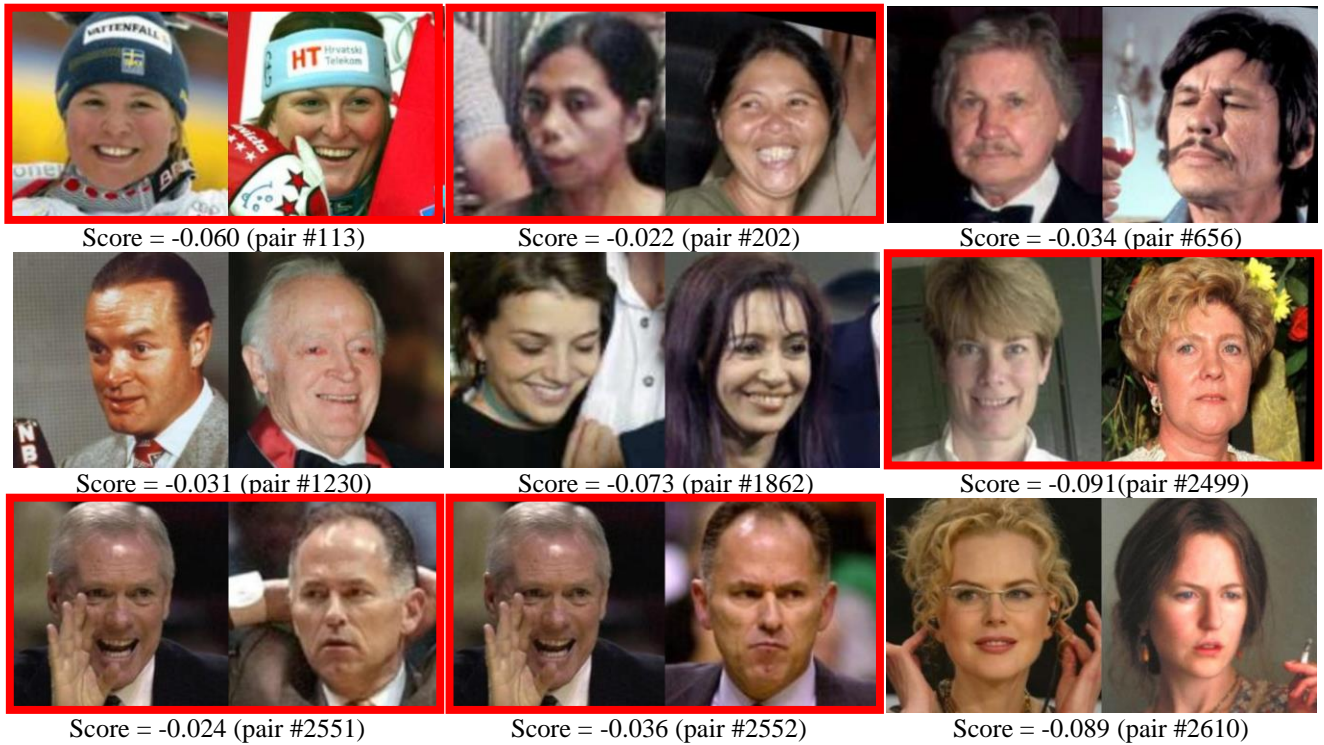

Score = -0.060 (pair #113)　　Score = -0.022 (pair #202)　　Score = -0.034 (pair #656)

Score = -0.031 (pair #1230)　　Score = -0.073 (pair #1862)　　Score = -0.091(pair #2499)

Score = -0.024 (pair #2551)　　Score = -0.036 (pair #2552)　　Score = -0.089 (pair #2610)

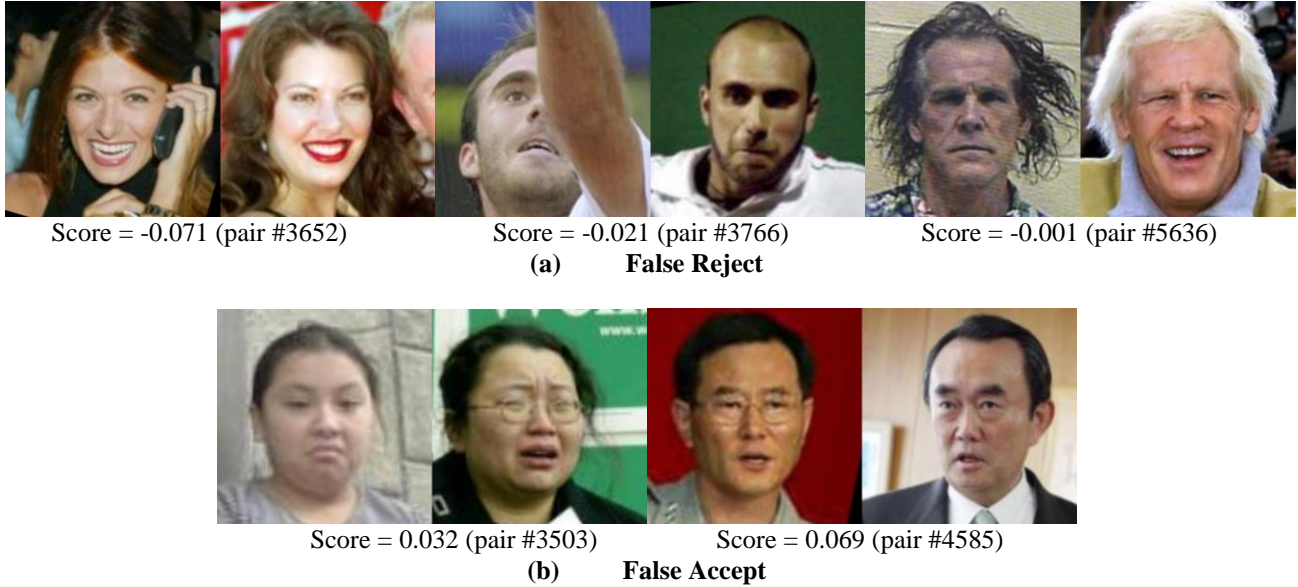

**Figure 3.** Failed cases in the LFW pair-wise verification task(including cases with wrong label): (a) False Reject Pairs. (b) False Accept Pairs. The score under the image is the similarity score of the above pair, and the score is in the range of [-1.0, 1.0] with threshold 0.0. The 5 pairs with red rectangle are wrong labeled ones.

## 4. DISCUSSION

As we know, face verification and open-set identification are the most usual applications of face recognition. For verification task, recall of our approach achieved 99.41% when the false alarm rate is 0.001, and even when the false alarm rate is 0.0001, the recall is 97.38%. It shows that, face verification performance is good enough to satisfy the needs of real applications. But for open-set identification, when the false alarm rate is 0.0001, the recall is about 80%. Although this is the best result in this task which is very promising, considering it is more prone to false alarm in identification scenario, we believe that the performance of open-set identification is still not good enough to satisfy the need of real applications.

We found that training data is very important For the performance of face recognition. We collected an evaluation dataset by mobile phone camera, which includes about 3300 Chinese people, and all the faces of one person are collected at different times. We use our model trained by celebrities to test the verification task on the evaluation dataset, and we achieved 85% when the false alarm rate is 0.0001. After we added Chinese celebrities collected from websites to train model with same parameters, we achieved 92.5% when the false alarm rate is 0.0001. We believe if we add more faces which are collected in the same situation as the evaluation dataset, we can achieve better results. We believe that data is as important as algorithm, and we suggest that before we can collect large amount of data in real situation, it is better not to draw conclusion that the face recognition approach is not good enough.

LFW has been the most popular evaluation benchmark for face recognition, and played a very important role in facilitating the face recognition society to improve algorithm. But after there are only 9 wrong pairs left(except the wrong labeled ones), which might reach the ultimate performance on LFW dataset, a new benchmark is expected to compare different approaches more effectively.

## 5. CONCLUSION

We propose a two-stage method for face recognition that combines deep CNN and metric learning. Benefited from features from multi-patch, our method can handle the cases with variant poses, occlusions and expressions well. As the amount of identities and faces per identity in training data increase, the performance improves correspondingly.

The proposed method outperforms state-of-the-art methods on LFW under main protocols and gains a quite high verification rate when the FAR is rather low. As the algorithm will keep improving, we hope that face recognition technique can eventually be widely used in more challenging conditions in the real world.

ACKNOWLEDGEMENT

We would like to thank Prof. Erik G. Learned-Miller for his responsive feedback and helpful advice on this paper.

REFERENCES

[1] Y. Taigman, M. Yang, M. Ranzato, and L. Wolf. DeepFace: Closing the gap to human-level performance in face verification. In Proc. CVPR, 2014.

[2] Erjin Zhou, Zhimin Cao, Qi yin. Naive-Deep Face Recognition: Touching the Limit of LFW Benchmark or Not? Technical report, arXiv:1501.04690.

[3] L. Best-Rowden, H. Han, C. Otto, B. Klare, and A. K. Jain. Unconstrained face recognition: Identifying a person of interest from a media collection. TR MSU-CSE-14-1, 2014.

[4] D. Yi, Z. Lei, S. Liao, and S. Z. Li. Learning face representation from scratch. In arXiv:1411.7923, 2014.

[5] G. B. Huang, M. Ramesh, T. Berg, and E. Learned-Miller. Labeled faces in the wild: A database for studying face recognition in unconstrained environments. Technical Report 07-49, University of Massachusetts, Amherst, October 2007.

[6] Y. Sun, Y. Chen, X. Wang, and X. Tang. Deep learning face representation by joint identification-verification. In Advances in Neural Information Processing Systems, pages 1988–1996, 2014.

[7] Y. Sun, X.Wang, and X. Tang. Deep learning face representation from predicting 10,000 classes. In Computer Vision and Pattern Recognition (CVPR), 2014 IEEE Conference on, pages 1891–1898. IEEE, 2014.

[8] Y. Sun, X. Wang, and X. Tang. Deeply learned face representations are sparse, selective, and robust. arXiv preprint arXiv:1412.1265, 2014.

[9] Yi Sun, Ding Liang, Xiaogang Wang, and Xiaoou Tang. DeepID3: Face Recognition with Very Deep Neural Networks. *arXiv:1502.00873*, 2014.

[10] D. Chen, X. Cao, F. Wen, and J. Sun. Blessing of dimensionality: Highdimensional feature and its efficient compression for face verification. In Computer Vision and Pattern Recognition (CVPR), 2013 IEEE Conference on, pages 3025–3032. IEEE, 2013.

[11] Xudong Cao, David Wipf, Fang Wen, and Genquan Duan. A Practical Transfer Learning Algorithm for Face Verification. International Conference on Computer Vision (ICCV), 2013.

[12] F. Schroff, D. Kalenichenko, and J. Philbin. Facenet: A unified embedding for face recognition and clustering. CVPR, 2015.

[13] Gary B. Huang and Erik Learned-Miller. Labeled Faces in the Wild: Updates and New Reporting Procedures. Technique report, University of Massachusetts , 2015.

[14] S. Liao, Z. Lei, D. Yi, and S. Z. Li. "A benchmark study of large-scale unconstrained face recognition". In IAPR/IEEE International Joint Conference on Biometrics, Clearwater, Florida, USA, 2014.

[15] Yaniv Taigman, Ming Yang, Marc'Aurelio Ranzato, Lior Wolf. "Web-Scale Training for Face Identification". CVPR, 2015.